\documentclass[10pt,twocolumn,letterpaper]{article}

\usepackage[pagenumbers]{cvpr} %

\usepackage{graphicx}
\usepackage{amsmath}
\usepackage{amssymb}
\usepackage{booktabs}

\usepackage{multirow}
\usepackage{adjustbox}

\usepackage{amsfonts}
\usepackage{pifont}
\newcommand{\cmark}{\ding{51}}%

\usepackage[ruled, vlined, noend]{algorithm2e}

\usepackage{mathtools} %

\makeatletter
\@namedef{ver@everyshi.sty}{}
\makeatother
\usepackage{tikz}
\usetikzlibrary{decorations.pathreplacing,positioning, arrows.meta}
\newcommand{\TimelineWidth}{11cm}

\newcommand{\method}{FedSpace\xspace}

\usepackage[accsupp]{axessibility}

\usepackage{xcolor}
\definecolor{ForestGreen}{RGB}{34,139,34}

\usepackage[pagebackref,breaklinks,colorlinks]{hyperref}

\usepackage[capitalize]{cleveref}
\crefname{section}{Sec.}{Secs.}
\Crefname{section}{Section}{Sections}
\Crefname{table}{Table}{Tables}
\crefname{table}{Tab.}{Tabs.}

\def\cvprPaperID{12} %
\def\confName{CVPR}
\def\confYear{2023}

\begin{document}

\title{Asynchronous Federated Continual Learning}

\author{Donald Shenaj, Marco Toldo, Alberto Rigon, Pietro Zanuttigh\\
University of Padova, Italy\\
}
\maketitle

\begin{abstract}
The standard class-incremental continual learning setting assumes a set of tasks seen one after the other in a fixed and pre-defined order. This is not very realistic in federated learning environments where each client works independently in an asynchronous manner getting data for the different tasks in time-frames and orders totally uncorrelated with the other ones.
We introduce a novel federated learning setting (AFCL) where the continual learning of multiple tasks happens at each client with different orderings and in asynchronous time slots.
We tackle this novel task using prototype-based learning, a representation loss, fractal pre-training, and a modified aggregation policy.
Our approach, called \method, effectively tackles this task as shown by the results on the CIFAR-100 dataset using 3 different federated splits with 50, 100, and 500 clients, respectively.
The code and federated splits are available at
\url{https://github.com/LTTM/FedSpace}.

\end{abstract}

\section{Introduction}
\label{sec:intro}
Federated Learning (FL) involves distributed learning across multiple heterogeneous devices, coordinated by a central server, without sharing private data.
In its standard fashion, FL implies a static configuration, where clients have access to the same local data and do not change their tasks for the entire training process. Furthermore, the training procedure is typically divided into rounds perfectly aligned across the different clients.

This setting, even if reasonable for research purposes, does not represent a realistic real-world learning setting where the clients (e.g., smartphones, autonomous cars, etc..) typically acquire data from onboard cameras or other devices while the training goes on and can be required to learn new tasks during time. Furthermore, each client asynchronously works independently from the others and can get new data at any time typically misaligned with the others.

Recently the continual learning (CL) paradigm has emerged as a more practical setting to mimic a real-world learning process, where new data and tasks are likely to be encountered over time.
We, therefore introduce a new setting combining continual and federated learning. We allow each client to follow its own continual learning path in an asynchronous way with respect to the other clients, while the server orchestrates the distributed training in order to safely aggregate the different learned models coming from the streams of information locally acquired at the clients.

\begin{figure}[tbp]
    \centering
    \includegraphics[width=0.5\textwidth]{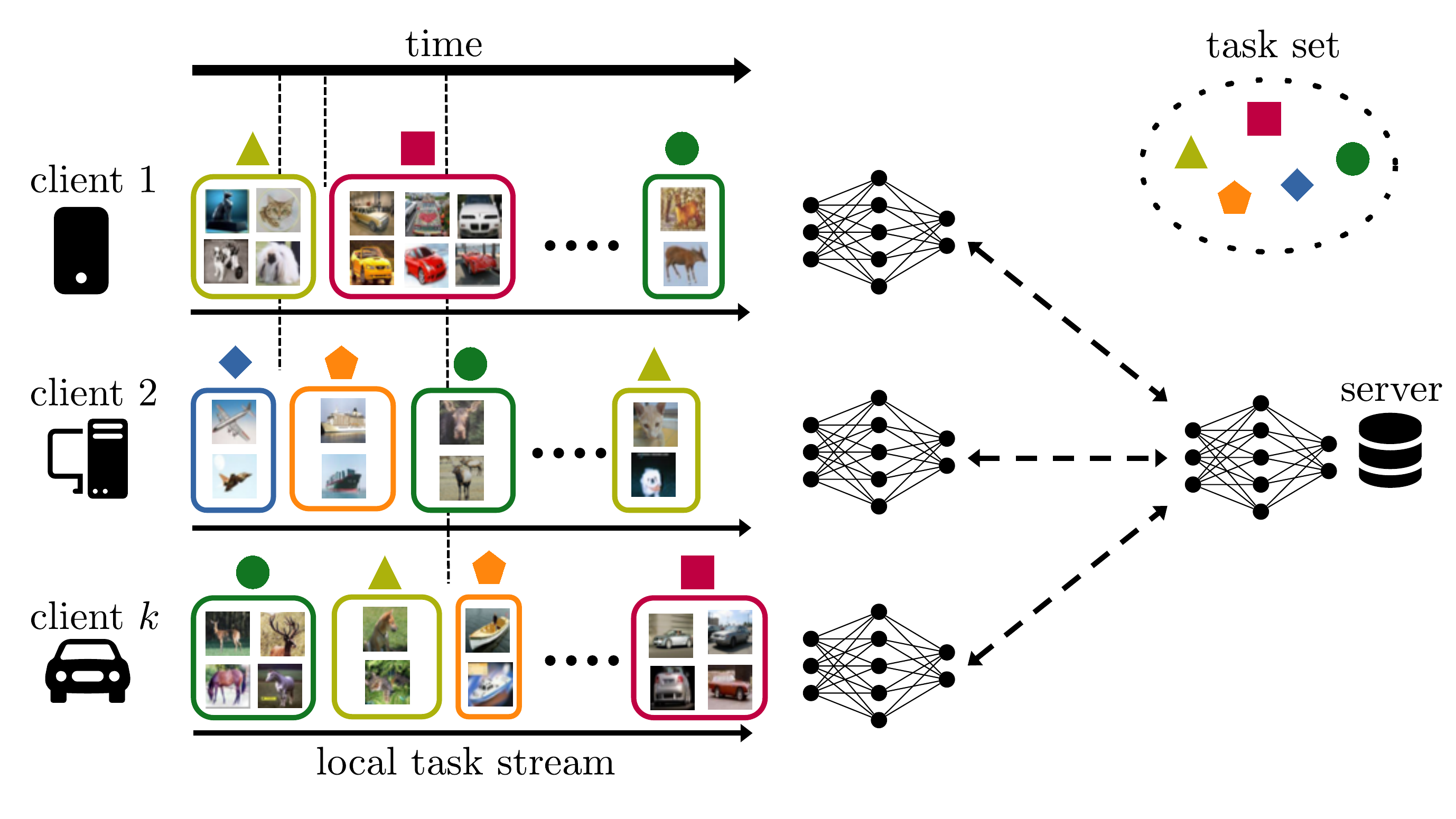}
    \caption{Illustration of the considered asynchronous federated continual learning setting (AFCL). Each client follows his own task stream, experiencing the global set of available tasks in different time slots. At the end of each round, the server  aggregates clients trained on  heterogeneous sets of classes.}
    \label{fig:setup}
    \vspace{0.5cm}
\end{figure}

More in detail, we propose to tackle a comprehensive setting, that we called Asynchronous Federated Continual Learning (AFCL),  in which each client of a federated system undergoes its own continual learning procedure, progressing task by task in an independent fashion, \ie, by:
\begin{itemize}
    \item following its individual sequence of tasks, which can be different from client to client,
    \item under its own temporal scheduling, i.e., the global training comprises out-of-sync local streams of data corresponding to different tasks.
\end{itemize}
The overall objective of AFCL is to end up with a prediction model that has assimilated all the tasks' knowledge acquired by  individual clients, avoiding catastrophic forgetting inherent in the incremental local task progression.

We identify the main challenges of the AFCL setting in:
\begin{itemize}
    \item \textbf{Asynchronous Learning}: Clients see different tasks at the same time, and switch between tasks in a misaligned manner.
    \item \textbf{Global CL}: Even though globally multiple tasks are experienced simultaneously, there will be tasks not currently observed, so it is crucial to preserve the information related to them.
\end{itemize}

To address this task we resort to a Federated learning System with Prototype Aggregation for Continual rEpresentation (abbreviated as \method), 
which exploits class prototypes in  feature space and contrastive learning both  to preserve previous knowledge and to avoid too divergent behavior between the different clients.

The main contributions of this paper are the following:
\begin{itemize}
    \item We address AFCL, a novel federated learning setting where clients explore different tasks in time-frames and orders totally uncorrelated with the other clients.
    \item We propose 3 realistic federated splits for the CIFAR-100 dataset, using 50, 100, and 500 clients.
    \item We evaluate the effectiveness of pre-training in federated learning using fractal images.
    \item We introduce an aggregation strategy for the prototypes and resort to prototype augmentation loss to tackle catastrophic forgetting, as done in  continual learning  \cite{michieli2021sdr}.
    \item We design a contrastive representation loss able to align the old aggregated prototypes coming from other clients, with their new locally learned representation while disentangling it from different classes.
    \item We adopt a modified model aggregation strategy that combines the client models with the previous server model to keep under control the model drift in this very challenging setting.
\end{itemize}
\vspace{0.3cm}

\section{Related Works}
\label{sec:related}
\textbf{Federated Learning}.
Federated Learning (FL) is a machine learning paradigm introduced in \cite{fedavg} as an alternative way to train a global model 
from a federation of devices  keeping their data local, 
and communicating to the server only the model parameters.
The iterative FedAvg  algorithm  \cite{fedavg} represents the standard approach to address FL. First, the server randomly selects  a subset of clients, which will perform training on their local data and send back to the server their updated models. Then, the server computes the global model as the weighted mean of the received models and repeats the previous step for many rounds. Despite its simplicity, FedAvg empirically demonstrates remarkable results.
In the context of systems heterogeneity, FedAvg does not allow participating devices to perform variable amounts of local work
based on the constraints of their underlying system.

This issue is addressed by FedProx \cite{fedprox}, which can be seen as a generalization and re-parametrization of FedAvg, providing a theoretical convergence guarantee when learning over data from non-identical distributions.

SCAFFOLD \cite{scaffold} can be seen as an
improved version of the distributed optimization algorithm \cite{shamir2014communication} where a fixed number of
(stochastic) gradient steps are used in place of a proximal point update. It deals with the problem of client drift by introducing control variates. In particular, the difference between the local control variate and the global control variate is used to correct the gradients in local training.

However, both FedProx and SCAFFOLD, do not provide advantages over FedAvg
when training deep learning models on image datasets. This motivates MOON \cite{li2021model} to propose a new approach for handling non-IID image datasets. It proposes to enforce similarity between model representations to correct the local training of individual parties,
 conducting contrastive learning at model-level.

Some works \cite{xu2021asynchronous,xie2019asynchronous} consider an asynchronous setting where each heterogeneous client sends the model update  as soon as the training  is finished and the aggregation at the server side is done using information from a single client at a time. %
This setup differs from our setting since in our case the asynchronously is not in the time slot but in the task progression and update while we perform the aggregation with multiple clients at a time as in the standard setting.

\textbf{Continual Learning}.
Deep neural networks experience catastrophic forgetting when trained on new data \cite{FRENCH1999128}.
The objective of continual learning, therefore,
is to enable a model to learn from a never-ending stream of data without the need to retrain the model from scratch every time new data become available. 

For practical computer vision applications, it has been extensively studied in a class incremental fashion \cite{de2021continual,parisi2019continual}.
A first direction to mitigate catastrophic forgetting is to store and replay samples of old classes  \cite{rebuffi2017icarl, castro2018end, wu2019large, douillard2020podnet}.
Another direction is to build a separate generative model able to reproduce old samples, and reuse them while training the model on new classes \cite{shin2017continual, wu2018memory,maracani2021recall}.

Prototype-based learning has also been widely used for continual learning \cite{zhu2021prototype,toldo2022bring,michieli2021sdr}. In PASS \cite{zhu2021prototype} a simple non-exemplar based technique is proposed to address the catastrophic forgetting problem in incremental learning. Instead of memorizing raw images, they memorize one class-representative prototype for each old class and adopt prototype augmentation by adding Gaussian noise in the  feature space to maintain the decision boundary of previous tasks.

\textbf{Federated Continual Learning}.
Very recently, some works started to address federated learning in settings, where the learned tasks change over time \cite{yao2020continual, yoon2021federated, dong2022federated, huang2022learn}, 
and where the clients' models are adapted to different domains \cite{yao2022federated, shenaj2023LADD}. 
Focusing on settings where the task changes over time, i.e., on continual learning,  some recent interesting works can be pointed out.

A preliminary work
\cite{yao2020continual} exploits the storage of training samples at the server side, compromising the privacy of the clients' data. In FedWEIT\cite{yoon2021federated} the network weights are decomposed into global federated parameters and sparse task-specific parameters. Each client receives selective knowledge from other clients by taking a weighted combination of their
task-specific parameters. However, they still require a buffer for data replay and does not scale when increasing the number of clients or tasks, due to the communication overhead.
The task of Federated Class-Incremental Learning is addressed in \cite{dong2022federated, dong2023no} with a focus on the privacy of the systems: differently from previous works, they only store perturbed images in a replay memory, while the server has no prior knowledge about when local clients will receive the data for new classes.

Another federated continual learning scenario is considered in \cite{Hendryx2021FederatedRE}.  Here, the server is pre-trained on a set of classes, corresponding to the first task. Then, each client learns the remaining ones by following its own task stream. Differently from the proposed AFCL setting, clients explore new sets of classes synchronously.

FCCL \cite{huang2022learn} 
deals with catastrophic forgetting and heterogeneity problems in federated learning, 
by leveraging unlabeled public data. In their setting,  ``continual'' refers to the fact that clients continually learn from each other as the training proceeds.

\section{Problem Formulation}

Let us consider a federated learning setting where $N$ clients tackle a classification problem aiming at assigning each data sample (\ie, an image in the considered setting) to one of the possible classes $c \in C$. We assume to use a generic deep learning model $\mathcal{M}=D \circ E$, composed of an encoder $E$ followed by a decoder $D$ as most architectures for image classification. 

\begin{figure*}
\centering
\begin{tikzpicture}

\node  (nodeXi) at (-1,0){client 1};
\draw[thick, -Triangle] (0,0) -- (\TimelineWidth,0) node[font=\scriptsize,below left=3pt and -8pt]{rounds $t$};

\foreach \x in {0,1,...,10}
\draw (\x cm,3pt) -- (\x cm,-3pt);

\foreach \x/\descr in {0/0,2/T_1,5/T_2,8/T_3}
\node[font=\scriptsize, text height=1.75ex,
text depth=.5ex] at (\x,-.3) {$\descr$};

\draw[orange!40, line width=4pt] (0,.5)-- +(2,0);
\draw[cyan!40, line width=4pt] (2,.5)-- +(3,0);
\draw[purple!40, line width=4pt] (5,.5)-- +(3,0);

\end{tikzpicture}

\begin{tikzpicture}
\node  (nodeXi) at (-1,0){client 2};
\draw[thick, -Triangle] (0,0) -- (\TimelineWidth,0) node[font=\scriptsize,below left=3pt and -8pt]{rounds $t$};

\foreach \x in {0,1,...,10}
\draw (\x cm,3pt) -- (\x cm,-3pt);

\foreach \x/\descr in {0/0,3/T_1,4/T_2,7/T_3}
\node[font=\scriptsize, text height=1.75ex,
text depth=.5ex] at (\x,-.3) {$\descr$};

\draw[blue!20, line width=4pt] (0,.5)-- +(3,0);
\draw[green!40, line width=4pt] (3,.5)-- +(1,0);
\draw[orange!40, line width=4pt] (4,.5)-- +(3,0);
\end{tikzpicture} 

\begin{tikzpicture}
\node  (nodeXi) at (-1,0){client $k$};
\draw[thick, -Triangle] (0,0) -- (\TimelineWidth,0) node[font=\scriptsize,below left=3pt and -8pt]{rounds $t$};

\foreach \x in {0,1,...,10}
\draw (\x cm,3pt) -- (\x cm,-3pt);

\foreach \x/\descr in {0/0,1/T_1,4/T_2,6/T_3, 9/T_4}
\node[font=\scriptsize, text height=1.75ex,
text depth=.5ex] at (\x,-.3) {$\descr$};

\draw[purple!40, line width=4pt] (0,.5)-- +(1,0);
\draw[cyan!40, line width=4pt] (1,.5)-- +(3,0);
\draw[blue!20, line width=4pt] (4,.5)-- +(2,0);
\draw[orange!40, line width=4pt] (6,.5)-- +(3,0);

\draw[line width=0.4pt] (5,1) circle (1.5pt);
\draw[line width=0.4pt] (5,1.3) circle (1.5pt);
\draw[line width=0.4pt] (5,1.6) circle (1.5pt);

\end{tikzpicture}

\caption{Example timeline showing the progression of the different tasks on the various clients. (\textit{Best viewed in colors)}}
\label{fig:rounds}
\end{figure*}
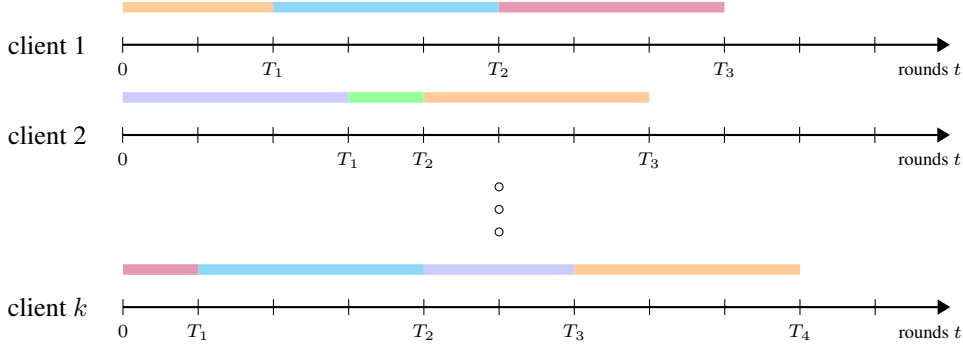

In the considered setting each client at each time step has access to a different subset of the data: let $C_k^{(t)} \subset C$ denote the set of classes (\ie, the task) available for client $k$ at round $t$. We denote with $\theta^{(t)} = \{\xi^{(t)}, \psi^{(t)}\}$ the parameters of the encoder and decoder models at round $t$. 
We assume that the client has access to the same data stream for a certain time interval of variable length, then moves to a new set of data, and so on.
We can thus define an ordered set of boundaries $T_k = \{T_{k,i}\}_{i=1}^{R_k}$, such that each client changes task only at the boundaries location, i.e., $C_k^{(t)}=C_k^{(T_{k,i})}  \quad \forall t: T_{k,i-1} < t \leq T_{k,i}$. $R_k$ denotes the total number of data streams for client $k$ (see Fig. \ref{fig:rounds}). 

Each client has a local stream dataset $\mathcal{D}=\{\mathcal{D}_k^{(T_{k, i})}\}_{i=1}^{T}$, where $\mathcal{D}_k^t=\{\mathbf{X} ,
\mathbf{Y}\}$ corresponds to the image-label pair. Note how the available data change at each boundary $T_{k,i}$ as depicted in Fig. \ref{fig:rounds}. Each client performs local training on its sequence of tasks $\{C_k^{(T_{k, i})}\}_{i=1}^{R_k}$ in an asynchronous way w.r.t the others (see Fig.~\ref{fig:setup} and Fig.~\ref{fig:rounds}).

At each communication round $t$, a subset of clients $K_t \subset N$ is randomly selected. Every client $k \in K_t$ downloads the global model (i.e., the model weights $\theta^{(t-1)}$), performs training on his local data $\mathcal{D}_k^{(t)}$,  and sends the updated weights $\theta_k^{(t)}$ to the server, which will perform aggregation and produce the new global model  $\theta^{(t)}$.

\section{Method}
\label{sec:method}
In this section, we are going to describe in detail the proposed method (\method) to tackle AFCL. Firstly the pre-training step at the server side exploiting fractal images  is discussed in 
Sec.~\ref{sec:pretrain}. Then, the optimization framework for each client is introduced in Sec.~\ref{sec:optimization}. The prototype aggregation and representation loss are detailed respectively in Sec.~\ref{sec:proto} and Sec.~\ref{sec:representation}.
Finally, the server aggregation is discussed in Sec.~\ref{sec:server}. 

\begin{figure*}
    \centering
    \includegraphics[width=0.9\textwidth]{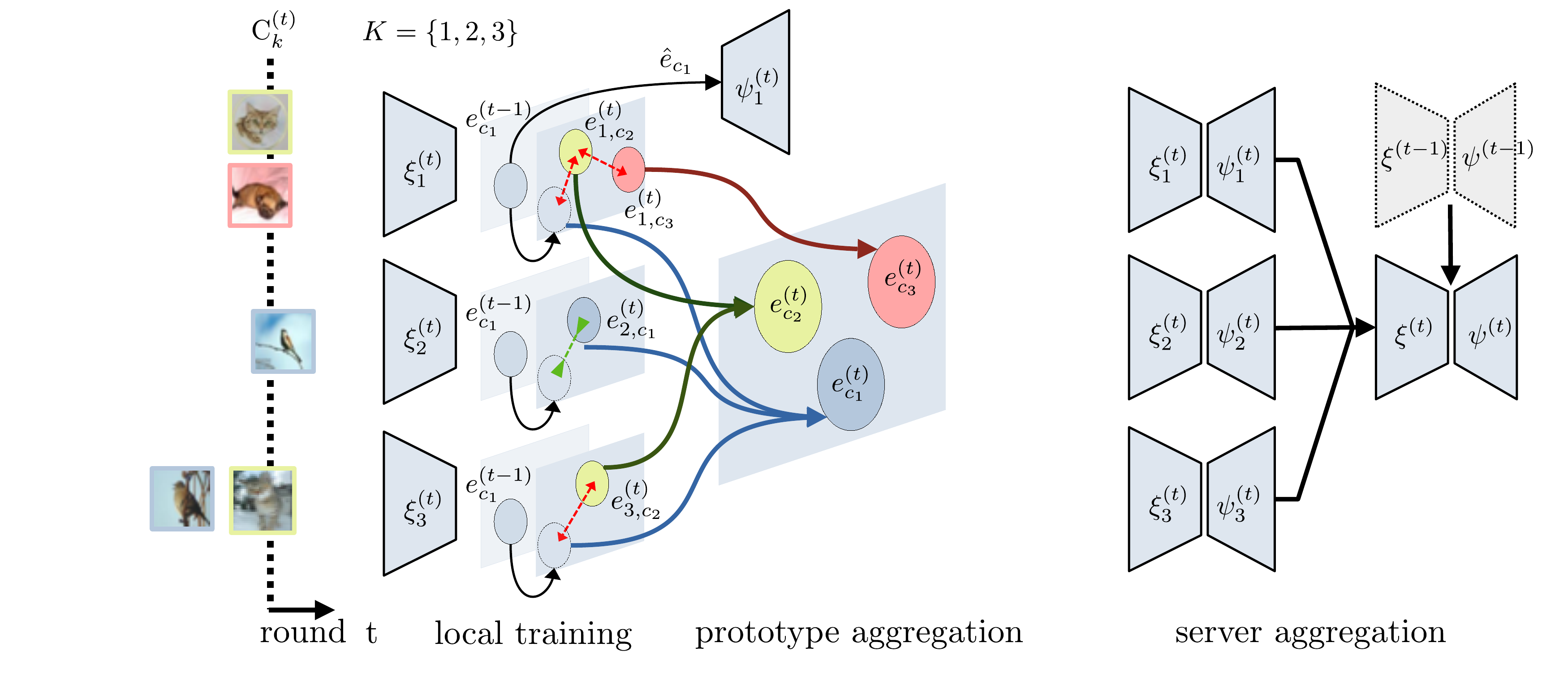}
    \caption{Overview of the \method learning algorithm, proposed for solving AFCL.}
    \label{fig:method}
\end{figure*}

The proposed framework is depicted in Fig.~\ref{fig:method} and summarized in Algorithm \ref{alg:method}. 

\begin{algorithm}[t]
\SetAlgoLined
\SetKwBlock{Pretraining}{Pre-training}{}
\SetKwBlock{Clients}{Federated training}{}
\textbf{Input:} Number of communication rounds T, \\
    $N$ clients each with its  data stream $\mathcal{D}=\{\mathcal{D}_k^{(T_{k, i})}\}_{i=1}^{R_k}$ \\    
\textbf{Output:} 
Model weights $\theta^{(T)}$\\
\Pretraining{
    Build a fractal dataset $D_f$\\ 
    Train the initial server model $\theta^{(0)}$ on $\mathcal{D}_f$
}
\Clients{
    $e_{k,c}^{(t)}, r_{k}^{(t)} = \emptyset, \emptyset$ \\
    \For{ round $t \in T$}{
        \For{client $k \in K$ in parallel}{
            Set $\theta_k^{(t)} = \theta^{(t-1)}$ \\
            $e_{k,c}^{(t)}, r_{k}^{(t)}\leftarrow$ \textsc{ComputeProto}(
            $\mathcal{D}_k^{(t)}$
            ) (Eq.~\ref{eq:compute_proto},  Eq.~\ref{eq:compute_radius}) \\
            Optimize $\theta_k^{(t)} $ with:
            $\mathcal{L} = \mathcal{L}_{CE} + \lambda_p \mathcal{L}_p +  \lambda_r \mathcal{L}_r  $ (Eq.~\ref{eq:client_optimization})
        }
        $\theta^{(t)} $ = \textsc{AggrModel}($\theta_k^{(t)}  \quad \forall k \in K$,$\theta^{(t-1)}$ ) (Eq.~\ref{eq:aggr_model}) \\
        $e_{c}^{(t)}, r^{(t)}\leftarrow$ \textsc{AggrProto}(
        $e_{k,c}^{(t)}, r_{k}^{(t)} \forall k\! \in K$)   (Eq.~\ref{eq:aggr_proto})
    }
}
\caption{\method learning scheme}
\label{alg:method}
\end{algorithm}

\subsection{Server Fractal Pre-training}
\label{sec:pretrain}
Recent studies suggest that pre-training allows to improve performances for the downstream task and speeds up training in federated learning \cite{chen2023importance, nguyen2022begin}. We resort to the pre-training strategy with fractal images  proposed in \cite{anderson2022improving} for a centralized setting, and use it to pre-train the server model before starting the learning at clients' side.
Fractal images are generated from affine Iterated Function Systems (affine IFS) \cite{kataoka2020pre}, therefore they provide perfect label accuracy at zero cost, no privacy issues, and most importantly do not contain images or classes present in the CIFAR-100 dataset, hence they are the best candidate in FL.
Additionally, in AFCL, since clients experience new classes inside different time slots, pretraining allows to limit client drift and  achieve a better alignment of their internal representations since they all start from an initial common feature space.
Following \cite{anderson2022improving}, we 
consider a set of classes $C_f$ (with $|C_f|=1000$, that is higher than the number of classes $|C|$ used in our experiments) and
build a dataset $\mathcal{D}_f = \{X_f, Y_f \}$ where $x_f \in X_f$ is a fractal image generated from the $y_f \in Y_f$ IFS code or label. Hence, training is performed using the standard cross-entropy loss for multi-class classification.
At the end of the pre-training, the server parameters $\theta^{(0)} = \{\xi^{(0)}, \psi^{(0)}\}$ are sent to every client $k \in K$, which  will proceed to update their model.
Notice that the number of local classes at client $k$, $|C_k^{(0)}|$ is smaller than the number of classes $|C_f|$ used during pre-training. Hence, let $\psi_{nc}^{(0)}$ be the classifier weights $\forall c \notin  C_k^{(0)} \cap C_f $, we can divide the decoder parameters into the ones corresponding to the not currently considered classes $\psi_{nc}^{(0)}$ and all the others $\hat{\psi}^{(0)}$: 
\begin{equation}
    \psi^{(0)} = \{ \hat{\psi}^{(0)}, \psi_{nc}^{(0)}\},
\end{equation}
and initialize each client model 
$\theta_k^{0}$ as follows:
\begin{align}
    \{\xi^{(0)}_k, \psi^{(0)}_k\} & \leftarrow  \{\xi^{(0)}, \hat{\psi}^{(0)}\}.
\end{align}

\subsection{Clients' Optimization}
\label{sec:optimization}
Each client performs local training optimizing the following loss function:
\begin{equation}
  \mathcal{L} = \mathcal{L}_{CE} + \lambda_p \mathcal{L}_p +  \lambda_r \mathcal{L}_r  
  \label{eq:client_optimization}
  \end{equation}
where $\mathcal{L}_{CE}$ is the standard supervised cross-entropy loss used for classification on $\mathcal{D}_k^{(t)}$,  $\mathcal{L}_p$ is the prototype-based  loss %
and finally $\mathcal{L}_r$ corresponds to our contrastive representation loss. The   $\lambda_p$ and $\lambda_r$ hyper-parameters are used as weighting factors for the respective losses.

\subsection{Prototype Aggregation}
\label{sec:proto}
At each round $t$, every selected client $k \in K$, for each class $c \in C_k^{(t)}$,
computes the mean feature vector, i.e., prototype of the class:
\begin{equation}
e_{k,c}^{(t)} = \frac{1}{|\mathcal{D}_k^{(t)}|}\sum_{\forall c \in C_{k}^{(t)}} E (\mathbf{X_k^{(t)};\xi_k^{(t)}}),
\label{eq:compute_proto}
\end{equation}
where $E(\cdot)$ is output of the encoder with weights $\xi_k^{(t)}$ given $\mathbf{X_k^{(t)}}$ in input.
Each client has also a radius $r_k^{(t)}$ capturing the variance of the associated features \cite{zhu2021prototype}:
\begin{equation}
r_k^{(t)} = \sqrt{\frac{1}{|\mathcal{D}_k^{(t)}|} \sum_{\forall c \in C_{k}^{(t)}} \frac{Tr(\Sigma_c^{(t)})}{d}},
\label{eq:compute_radius}
\end{equation}
where $Tr(\cdot)$ denotes the trace operator of a matrix, $\Sigma_c^{(t)}$ is the covariance matrix for the features of class $c$ at round $t$ for client $k$, and $d$ is the dimension of the feature space.

Then, each client sends the prototype-radius pair $(e_{k,c}^{(t)}, r_{k}^{(t)})$ to the server which will proceed to aggregate them into global prototypes $(e_c^{(t)}, r^{(t)})$ by computing their weighted average.
When at the server side an older representation $e_c^{(t-1)}$ already existed for a specific class $c$ 
(\ie, at least one client $m$ selected at round $t_0 < t$ performed training with $c \in C_k^{(t_0)} $), it is updated with a moving average with weight $\beta$. Therefore, the prototype aggregation becomes:

\begin{equation}
    e_{c}^{(t)} =
\begin{dcases*}
\sum_{k=1}^K \frac{|\mathcal{D}_k|}{\mathcal{D}} e_{k,c}^{(t)}
   & if  $c\! \notin \! C^{(t-1)}$ \\[1ex]
\beta \!\left( \sum_{k=1}^K  \frac{|\mathcal{D}_k|}{\mathcal{D}} e_{k,c}^{(t)} \right) + (1\!-\!\beta)   e_{c}^{(t-1)}  
   & otherwise\,.
\end{dcases*}
\label{eq:aggr_proto}
\end{equation}

During local training, each client $k$ performs prototype augmentation  \cite{zhu2021prototype} on the global prototypes $e_{c}^{(t-1)}$ of classes $c \notin \mathcal{C}_k^{(t)}$  not present in the current learning stage by adding Gaussian noise scaled by the radius $r^{(t-1)}$:
\begin{equation}
\hat{e}_c = e_c^{(t-1)} + r^{(t-1)} * \mathcal{N}(0,1)
\label{eq:proto_augmentation}
\end{equation}

Finally, a vector of augmented prototypes $\mathbf{\hat{e}}$ of the same length of the batch size is constructed (for each element $\mathbf{\hat{ e}}[n]$ its class $c=\mathcal{Y}[n]$ is randomly selected among the ones not present in the current stage).
The loss $\mathcal{L}_p$ is thus computed as the cross entropy between the prediction computed from the augmented prototypes and their labels:
\begin{equation}
\mathcal{L}_p = \sum_{n}%
\mathcal{L}_{CE} ( D(\mathbf{\hat{e}}[n]; \psi_k^{(t)}), \mathcal{Y}[n] ).
\end{equation}

\subsection{Representation Loss}
\label{sec:representation}
Previous works showed that contrastive learning can be efficiently used in continual learning to tackle catastrophic forgetting \cite{michieli2021sdr,cha21co2l,wang23survey}. Thus, we introduce an additional  loss function based on a contrastive paradigm %
that encourages feature vectors of the same class
to be close together %
while pushing away from each other 
feature vectors belonging to different classes.
At round $t$, each %
non-empty global prototype $e_i^{(t-1)}$ corresponding to a generic class $i \in C^{(t-1)} \setminus  \mathcal{C}_k^{(t)}$ is augmented according to Eq.~(\ref{eq:proto_augmentation}). Note that $C^{(t-1)} \setminus  \mathcal{C}_k^{(t)}$, $C^{(t-1)} \subset C$, represents the set of classes globally discovered by the server up to round $t-1$ but excluding the ones currently being learned. For these classes, a prototype has already been computed and an augmented prototype vector can be computed from Eq.~(\ref{eq:proto_augmentation}). Finally, the augmented prototype vectors and the  feature vectors computed on $\mathcal{C}_k^{(t)}$
are concatenated to form a new set of feature vectors.

For every  class $c \in \mathcal{C}_k^{(t)}$, a positive and a negative set of feature vectors are constructed. The positive set is represented by a matrix of pairwise cosine similarities between the feature vectors belonging to the considered class, both computed from input data or from augmented prototypes. The negative set is instead computed using feature vectors of all the other classes (again both from input or augmented prototypes).

Formally, the representation loss $\mathcal{L}_{r}$ is defined as:

\begin{equation}
\mathcal{L}_{r} \!=\! -\frac{1}{|\mathcal{D}_k^{(t)}|}\!\sum_{c=1}^{|\mathcal{C}_k^{(t)}|} \frac{1}{N_c(N_c\!-\!1)}\!\sum_{i\neq j}\log\frac{e^{s_{i,j}^+}}{e^{s_{i,j}^+}+\sum\limits_{k=1,k\neq c}^{C} e^{s_{i,k}^-}},
\label{eq:representation}
\end{equation}
where $N_c$ is the number of samples of the class $c$, $s_{i,j}^+$ is the cosine similarity between the $i$-th and $j$-th feature vectors in the positive set, $s_{i,k}^-$ is the cosine similarity between the $i$-th feature vector in the positive set and the $k$-th vector in the negative set.
The loss is computed as the negative log-likelihood of the positive set relative to the negative set averaged over the set of classes $\mathcal{C}_k^{(t)}$. 

The loss is minimized by adjusting the feature vectors to minimize the distances inside the positive set and maximize the distances between the positive set and the negative set elements.

In this way,
the loss enforces each client to learn locally new classes, clustering their features in a consistent way with their past representation obtained from the server (and therefore from the other clients), while at the same time %
enforcing separation between features of different classes.
This allows to obtain a feature clustering that is consistent across different clients and across time slots.

\subsection{Server  Aggregation}
\label{sec:server}
At the end of each communication round, each client sends the model to the server, where the aggregation is performed.
Typically in standard FedAvg \cite{fedavg} the server model is updated with the weighted average of the model parameters of each client participating in that round, by weighting his contribution depending on the number of images seen in that round. In our setting, there is no guarantee that all the classes are present in the considered step and furthermore some clients could have just started the learning with the new data stream and have a not very reliable model. In order to stabilize the aggregation avoiding drifts due to unreliable clients and the forgetting of old classes, we  perform a weighted average between the newly computed model and the previous model present at the server side. Mathematically:
\begin{equation}
    \mathbf{\theta}^{(t)} = \rho\left(\sum_{k=1}^K \frac{|\mathcal{D}_k|}{\sum_{i=1}^k|\mathcal{D}_i|} \mathbf{\theta}_k^{(t)}\right) + (1\!-\!\rho) \mathbf{\theta}^{(t-1)},
\label{eq:aggr_model}
\end{equation}
where $\mathbf{\theta}$ denotes the server model weights, $\mathbf{\theta}_k$ denotes the model weights of the client $k$, $|\mathcal{D}_k|$ is the number of images of client $k$, and $t$ corresponds to the communication round. We experimentally found that setting $\rho=0.5$, i.e., a simple average, provides the best performances.
\begin{figure*}[htbp]
\centering
\begin{subfigure}{.5\textwidth}
  \centering
    \includegraphics[width=\textwidth]{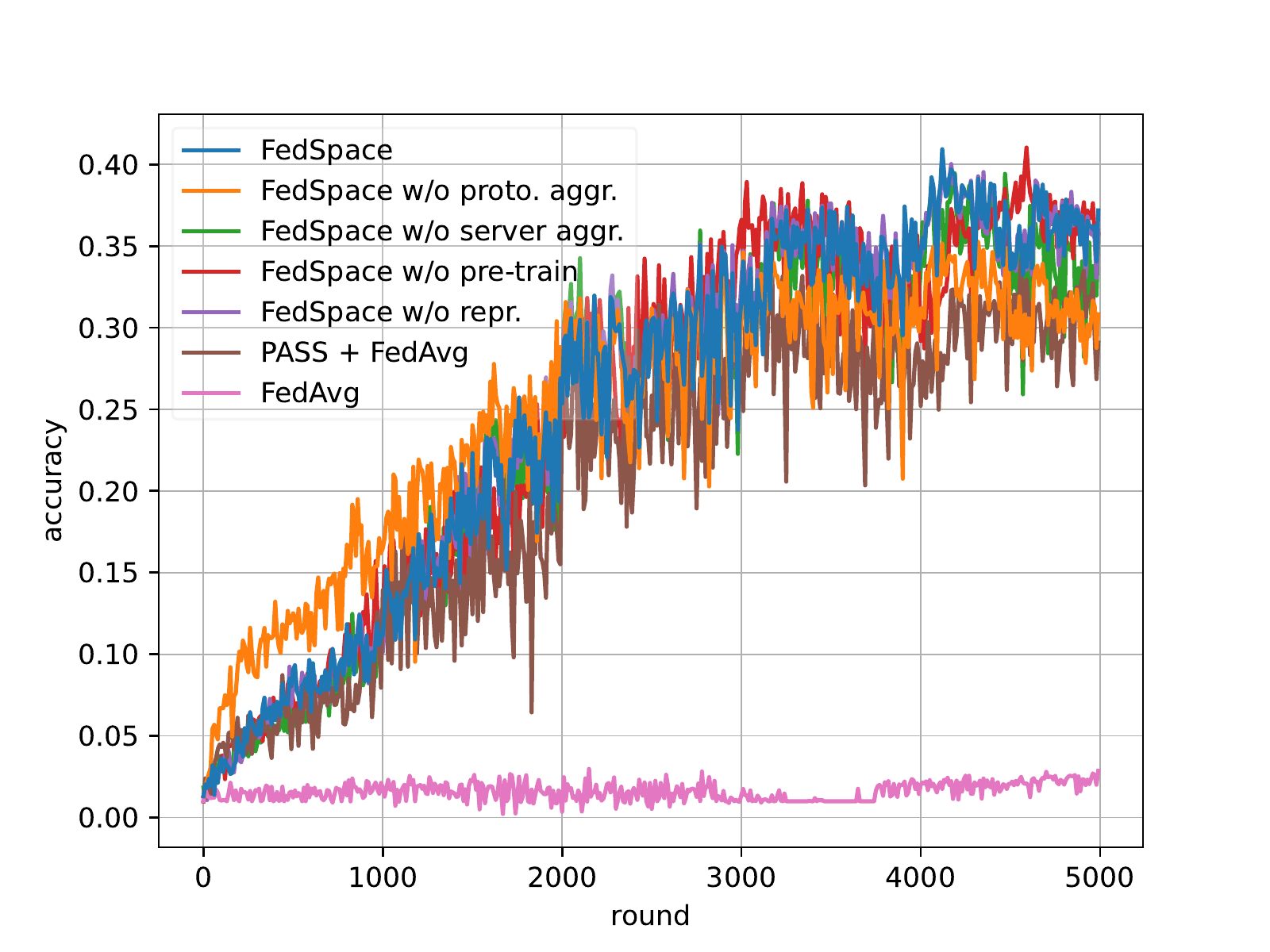}
    \caption{50 clients.}
    \label{fig:ablation_50_clients}
\end{subfigure}%
\begin{subfigure}{.5\textwidth}
  \centering
    \includegraphics[width=\textwidth]{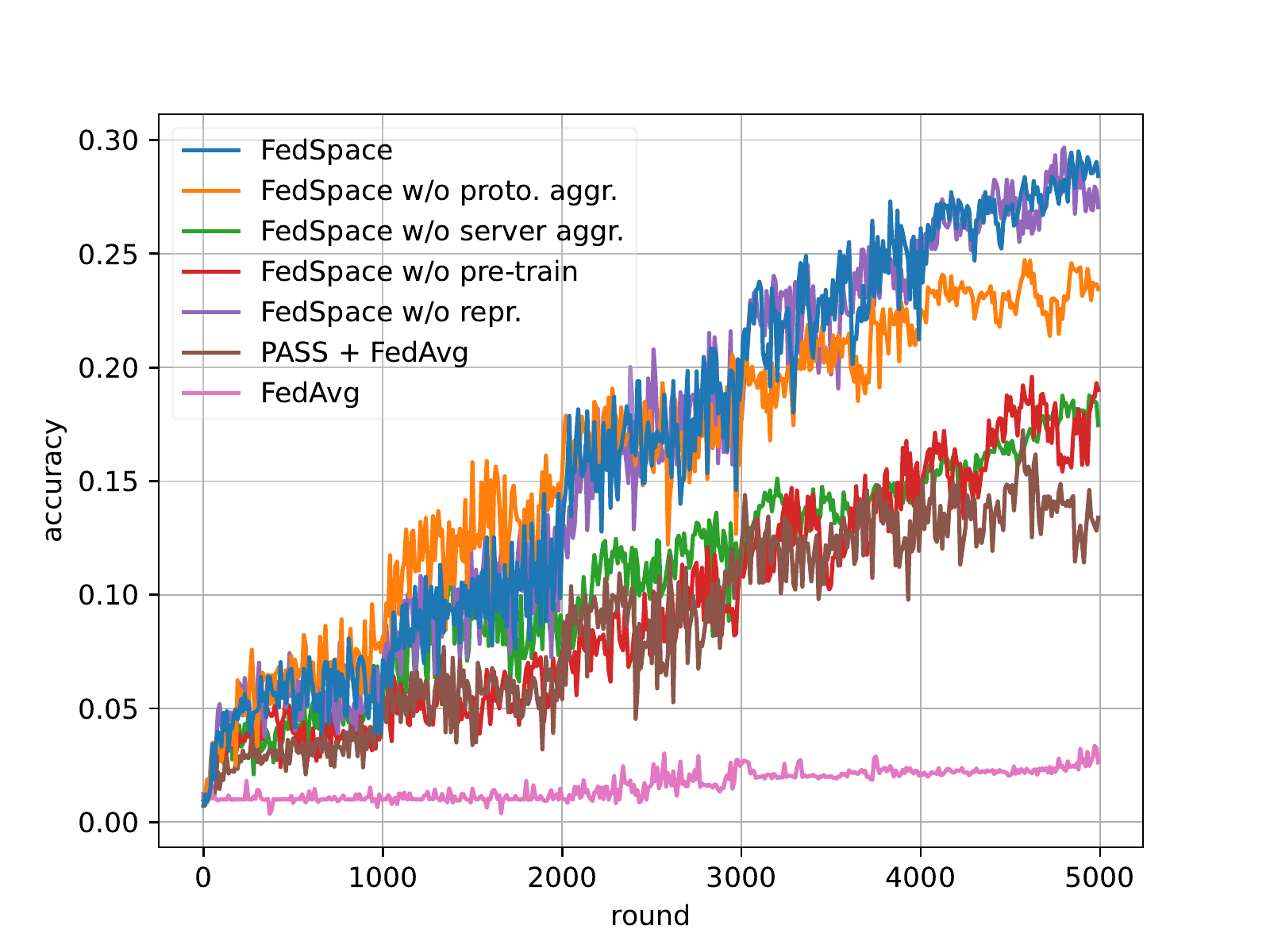}
    \caption{ 500 clients.}
    \label{fig:ablation_500_clients}
\end{subfigure}
\caption{Classification accuracy computed at every communication round. Refer to Table \ref{tab:ablation} for the notation.}
\label{fig:plot_ablation}
\vspace{0.7cm}
\end{figure*}

\section{Experiments}
\label{sec:results}

\subsection{Experimental Setup}
\textbf{Datasets}. To evaluate the performance of our method, we run the simulations on the CIFAR-100 dataset. 
\cite{krizhevsky2009learning}. We introduce 3 strongly non-IID splits with 10 tasks each made of 10 classes, while the total number of clients is equal to 50, 100, and 500, respectively.
To create them, first, a distribution of the number of samples for each client is computed using a power-law distribution. Next, for each client, the partitioning is done based on a Dirichlet distribution, where the proportions of samples from each class are randomly sampled from the distribution. The parameter of the Dirichlet distribution $\alpha$ was set equal to 3. 
Each client sets his task streams and boundaries (\ie, number of rounds per task) randomly and independently from other clients.

\textbf{Competitors}. Given the novelty of the setting, there is no other work explicitly developed for the task. However, we compared our proposed method with a baseline, e.g., FedAvg \cite{fedavg}, that is the standard approach for FL and with PASS \cite{zhu2021prototype}, a state-of-the-art non-exemplar algorithm for class incremental learning, purposefully adapted to the considered FL setting. In practice, it computes local prototypes at each round and makes use of a prototype-based loss. The  aggregation at the server side is done using FedAvg.

\textbf{Implementation details}.
The code is implemented in PyTorch, starting from 
the FedML \cite{he2020fedml} framework. 
The employed classification network  is ResNet-18 \cite{he2016deep}.
We run the experiments for 5000 communication rounds, $|K|=5 $ clients per round. The employed optimizer is ADAM, we used a batch size equal to 64, and a learning rate  $l_r=1e-3$, which is divided by 2 every 1000 rounds.
Concerning the parameters of our method we used $\beta=0.1$, $\lambda_{p} = 1e-2$, $\lambda_{r}=1e-2$. We also employed the self-supervised label augmentation of \cite{zhu2021prototype}.
We run the pre-training for one epoch with a batch size of 32, with images of size $256\times256$.
Training is performed sequentially on each client,  and run on a single  NVIDIA RTX 3090.
The code, federated splits, and configuration files are publicly available at
\url{https://github.com/LTTM/FedSpace}.

\subsection{Results}
In Table \ref{tab:comparison}, we report the
 Top-1 accuracy, for each federated split (50, 100, and 500 clients), and compare the performance with FedAvg \cite{fedavg} and PASS\cite{zhu2021prototype} adapted to FL.
We notice that in this highly challenging scenario, FedAvg \cite{fedavg} is not a feasible solution and provides very poor performances between $2\%$ and $3\%$ (even by adding the fractal pre-training stage of Section \ref{sec:pretrain} results do not improve). The main reason is the catastrophic forgetting, here experienced both 
in a continual learning and in  a federated learning way.
At each communication round, due to client drift  and misaligned learning of new classes across the selected clients, the global model loses knowledge of the previously learned classes.
Moreover, when a set of classes is no more experienced by clients selected in future rounds, those are forgotten as well known in continual learning. 

Since a continual learning scheme is fundamental to tackle this setting, we considered a state-of-the-art approach for this task, i.e, PASS \cite{zhu2021prototype} and combined it with the FedAvg aggregation scheme. We select this approach since it has impressive performances, does not use exemplar memory, and is based on prototypes as our approach.
Compared to FedAvg the performance improvement is large. With $50$ clients it achieves $29.1\%$, a reasonable score in this very challenging setting. When increasing the number of clients, however, the performance drop is significant, achieving around $23\%$ with $100$ clients and $13.4\%$ with $500$. Notice how especially in the latter case the drop is very relevant, showing that the approach has issues in scaling to a large number of clients.

Our approach achieves $37.2\%$ with $50$ clients, outperforming PASS of more than $8\%$. Furthermore with $100$ and $500$ clients it achieves $30\%$ and $28.4\%$, respectively. Especially the latter case shows how our approach scales much better with the number of clients more than doubling the performance of PASS in the most challenging setting.

\begin{table}[]
    \centering
    \begin{tabular}{l c c c }
            \hline
    \multirow{2}{*}{Method}&\multicolumn{3}{c}{\textbf{Top-1 Accuracy (\%)}}\\ 
         & $50$ & $100$ & $500$   \\
        \hline
        FedAvg \cite{fedavg} & 2.86 & 2.01 & 2.63\\
        PASS\cite{zhu2021prototype} + FedAvg \cite{fedavg} & 29.08 & 22.97 & 13.39 \\ 
        \textbf{Ours (\method)} & \textbf{37.18} & \textbf{29.99} & \textbf{28.42}\\
                \hline
    \end{tabular}
    \caption{Comparison of different methods for 50, 100, and 500 clients splits of CIFAR-100. Top-1 accuracy.}
    \label{tab:comparison}
\end{table}

\subsection{Ablation Studies }
In order to validate the performance and robustness of the proposed method, we perform an extensive ablation on all the three settings.  
In Table \ref{tab:ablation}, we report the Top-1 accuracy, computed on all classes $C$ after $5000$ rounds, while in Fig.~\ref{fig:plot_ablation} we show the classification accuracy while the training goes on, i.e.,  at the end of every communication round, for 50 and 500 clients.
The  Figure reports the accuracy during training for the competitors, our method, and its reduced versions obtained by removing each component singularly.

Interestingly, when the number of total clients is low, during the first rounds keeping the prototypes local (i.e., not aggregating them) leads to
higher performances, however, after a certain amount of rounds, aggregating prototypes starts to be beneficial and gives a similar improvement in every setting, with an increase of $6\%$, $4\%$ and $5\%$ for $50$, $100$ and $500$ clients, respectively (second row of Table \ref{tab:ablation}). Even if it is less evident, this is true also for the representation loss, with an increase of $1\%$, $3\%$ and $1\%$ (fourth row of Table \ref{tab:ablation}). 

When removing the modified server aggregation (\ie, performing aggregation as in FedAvg), the performance decrease by $1.5\%$ for $50$ clients, up to $11\%$ with $500$ clients. 
Similarly, by removing the pre-training step with $50$ clients, the drop is $1.5\%$, while when using $500$ clients it determines a larger decrease of $10\%$.

\begin{table}[htbp]
    \centering
    \begin{adjustbox}{width=\columnwidth}
    \centering
    \footnotesize
    \begin{tabular}{ccccccccc}
    \toprule
       {\textbf{Proto}}  & 
       {\textbf{Pre-}} & 
      {\textbf{Repr.}} & 
       {\textbf{Server}} &  
       \multicolumn{3}{c}{\textbf{Accuracy (\%)}} \\ 
      \textbf{ Aggr.} & \textbf{train}& \textbf{Loss} & \textbf{Aggr.}&  50 & 100 & 500 \\ 
        \midrule
            \cmark &  &  &   & 28.76 & 29.58 & 23.45   \\
             & \cmark & \cmark  &   \cmark & 30.82 & 25.89 & 23.45  \\
           \cmark & & \cmark  &   \cmark &  35.72 &  25.89 & 19.01 \\
           \cmark & \cmark &  &   \cmark & 35.96& 26.84 & 27.05  \\
            \cmark & \cmark & \cmark &   &  35.60 & 31.90 & 17.46  \\
           \cmark & \cmark & \cmark  &   \cmark & 37.18 & 29.99 & 28.42  \\
    \bottomrule
    \end{tabular}
    \end{adjustbox}
    \vspace{-5pt}
    \caption{Ablation of our method on CIFAR-100 for 50, 100 and 500 clients.
    Top-1 accuracy.
    Proto aggr. refers to the prototype aggregation (Eq.~\ref{eq:aggr_proto}), Pre-train is the server fractal pre-training (Sec~\ref{sec:pretrain}), Repr. Loss is the representation loss (Eq.~\ref{eq:representation}) and Server Aggr. corresponds to the modified aggregation (Eq.~\ref{eq:aggr_model}) instead of FedAvg.}
    \label{tab:ablation}
\end{table}

\section{Conclusions}
\label{sec:conclusions}

In this paper, we introduced a novel realistic setting for asynchronous federated continual learning (AFCL) detailing the main challenges.
Then, we presented our framework (FedSpace) which starts by 
performing pre-training at the server side with fractal images.
Next, local training on clients is driven by a loss applied to the augmented old global prototypes and by a contrastive representation loss used to properly allocate the prototypes of new (global) classes while at the same time aligning the new local classes with their past location. Finally, a modified server aggregation strategy reduces client drift.
To evaluate the performance of FedSpace, 
we proposed 3 federated splits of the CIFAR-100 dataset, using 50, 100, and 500 clients, that we used for the experimental evaluation.
Our approach reached state-of-the-art results when compared with competing methods adapted to our setting.

Further research will be devoted to the development of more advanced aggregation strategies at the server side and to the combination of the proposed approach with other state-of-the-art continual learning methods.

\section*{Acknowledgment}
This work was partially supported by the European Union under the Italian National Recovery and Resilience Plan (NRRP) of NextGenerationEU, partnership on “Telecommunications of the Future” (PE0000001 - program “RESTART”). 

{\small
\bibliographystyle{ieee_fullname}
\bibliography{egbib}
}

\end{document}